\documentclass[sigconf]{acmart}
\usepackage{bm}
\def\BibTeX{{\rm B\kern-.05em{\sc i\kern-.025em b}\kern-.08emT\kern-.1667em\lower.7ex\hbox{E}\kern-.125emX}}

\usepackage{stfloats}
\usepackage{epsfig}
\usepackage{graphicx}
\usepackage{pifont}
\usepackage{subfigure}
\usepackage{enumitem}
\usepackage{multirow}
\usepackage{array}

\usepackage{algorithm,algcompatible,amssymb,amsmath}

\algnewcommand\algorithmicto{\textbf{to}}
\algnewcommand\RETURN{\State \textbf{return} }
\algnewcommand\algorithmicinput{\textbf{Input:}}
\algnewcommand\INPUT{\item[\algorithmicinput]}
\algnewcommand\algorithmicoutput{\textbf{Output:}}
\algnewcommand\OUTPUT{\item[\algorithmicoutput]}
\algnewcommand\algorithmicinitialize{\textbf{Initialize:}}
\algnewcommand\INITIALIZE{\item[\algorithmicinitialize]}

\acmSubmissionID{698}

\makeatletter 
  \newcommand\figcaption{\def\@captype{figure}\caption} 
  \newcommand\tabcaption{\def\@captype{table}\caption} 
\makeatother

\newcommand{\ljx}[1]{\textcolor{black}{#1}}
\newcommand{\czq}[1]{\textcolor{black}{#1}}
\newcommand{\ljxa}[1]{\textcolor{black}{#1}}

\newcommand{\beq}{\begin{equation}}
\newcommand{\eeq}{\end{equation}}

\newcommand{\PP}{\mathbb{P}}

\newcommand{\NN}{\mathcal{N}}
\newcommand{\II}{\mathcal{I}}

\newcommand{\KL}[2]{D_{KL}(#1\mid\mid#2)}
\newcommand{\PJ}[2]{\mathbb{P}_{#1#2}}  
\newcommand{\PI}[2]{\mathbb{P}_{#1}\otimes\mathbb{P}_{#2}}  

\newcommand\Eq{Eqn.}
\newcommand\Eqs{Eqns.}
\newcommand\Fig{Figure}

\newcommand\Sec{Sec.}
\newcommand\Tbl{Table}
\newcommand\Alg{Alg.}

\copyrightyear{2019} 
\acmYear{2019} 
\acmConference[MM '19]{Proceedings of the 27th ACM International Conference on Multimedia}{October 21--25, 2019}{Nice, France}
\acmBooktitle{Proceedings of the 27th ACM International Conference on Multimedia (MM '19), October 21--25, 2019, Nice, France}
\acmPrice{15.00}
\acmDOI{10.1145/3343031.3350898}
\acmISBN{978-1-4503-6889-6/19/10}
\acmSubmissionID{698}

\begin{document}
\fancyhead{}

\title[]{Improving the Learning of Multi-column Convolutional Neural Network for Crowd Counting}

\author{Zhi-Qi Cheng$^{1,2*}$, Jun-Xiu Li$^{1,3*}$, Qi Dai$^3$, Xiao Wu$^{1\dagger}$, Jun-Yan He$^1$, Alexander G. Hauptmann$^2$}
\affiliation{\institution{$^1$Southwest Jiaotong University, $^2$Carnegie Mellon University, $^3$Microsoft Research}}
\email{{zhiqic, alex}@cs.cmu.edu, {lijunxiu@my, wuxiaohk@home}.swjtu.edu.cn, qid@microsoft.com, junyanhe1989@gmail.com}

\renewcommand{\shortauthors}{Zhi-Qi Cheng et al.}

\begin{abstract}
\label{sec:abstract}
Tremendous variation in the scale of people/head size is a critical problem for crowd counting.
To improve the scale invariance of feature representation, recent works extensively employ Convolutional Neural Networks with multi-column structures to handle different scales and resolutions.
However, due to the substantial redundant parameters in columns, \czq{existing} multi-column networks invariably exhibit almost the same scale features \czq{in different columns}, which severely affects counting accuracy and leads to overfitting.
In this paper, we attack this problem by proposing a novel Multi-column Mutual Learning (McML) strategy.
It has two main innovations:
1)~A statistical network is incorporated into the multi-column framework to estimate the mutual information between columns, which can \czq{approximately} indicate \czq{the scale correlation} between features from different columns.
By minimizing the mutual information, each column is guided to learn \ljxa{features} with different \czq{image scales}.
2)~We devise a mutual learning scheme that \czq{can alternately optimize each column} while keeping the other columns fixed \czq{on each mini-batch training data}.
With such asynchronous \czq{parameter update process}, each column is inclined to learn different feature representation from others, which can efficiently reduce the parameter redundancy and \czq{improve generalization ability}.
More remarkably, McML can be applied to all existing multi-column networks and is end-to-end trainable.
Extensive experiments on four challenging benchmarks show that McML can significantly improve the original multi-column networks and outperform the other state-of-the-art approaches.
\end{abstract}

\keywords{Crowd Counting; Multi-column Network; Mutual Learning Strategy}

\maketitle

{\fontsize{8pt}{8pt} \selectfont
	\textbf{ACM Reference Format:}\\
	Zhi-Qi Cheng, Jun-Xiu Li, Qi Dai, Xiao Wu, Jun-Yan He, Alexander G. Hauptmann. 2019. Improving the Learning of Multi-column Convolutional Neural Network for Crowd Counting. In \textit{ Proceedings of the 27th ACM International Conference on Multimedia (MM '19), Oct. 21--25, 2019, Nice, France.} 
	ACM, New York, NY, USA, 11 pages. https://doi.org/10.1145/3343031.3350898 }

\section{Introduction}
\label{sec:introduction}
\begin{figure}[!t]
\small
\begin{center}
\includegraphics[width=0.9\linewidth]{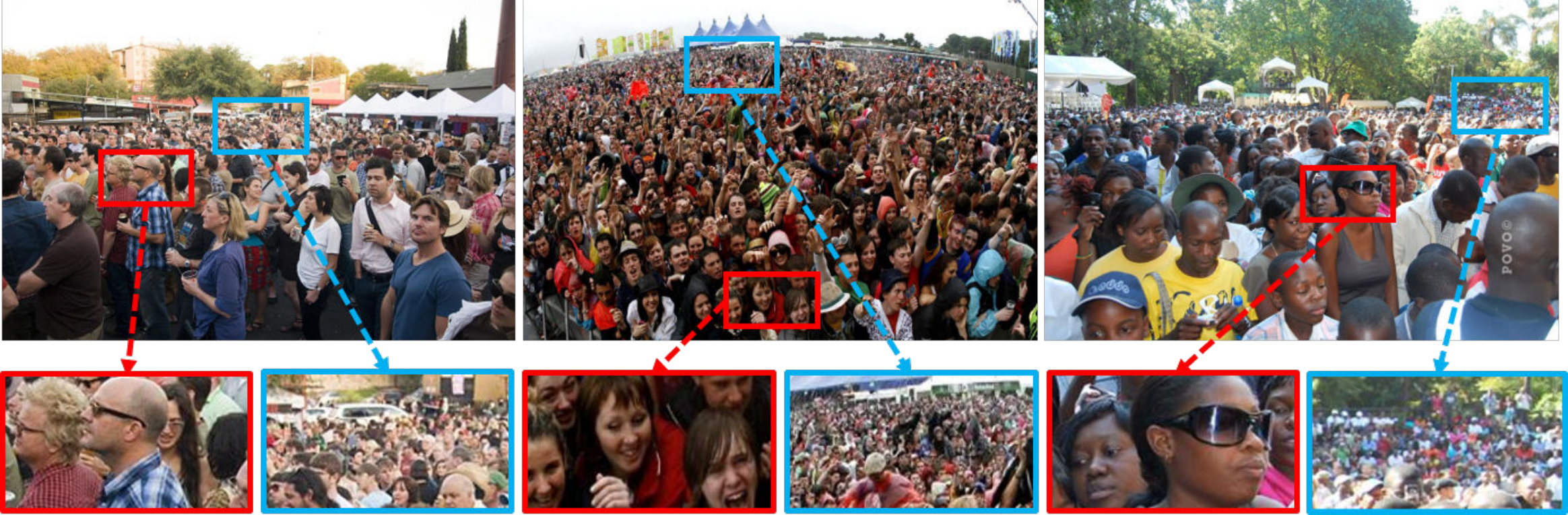}
\end{center}
\vspace{-4mm}
\caption{\small 
Examples of ShanghaiTech Part A dataset~\cite{MCNN-16}.
\ljx{Crowd counting is a challenging task with the significant variation in the people/head size due to the perspective effect.}
}
\label{fig:scale-changes}
\vspace{-4mm}
\end{figure}

With the growth of wide applications, such as safety monitoring, disaster management, and public space design, crowd counting has been extensively studied in the past decade.
As shown in Figure \ref{fig:scale-changes}, a significant challenge of crowd counting lies in the extreme variation in the scale of people/head size.
To improve the scale invariance of feature learning, Multi-column Convolutional Neural Networks are extensively studied~\cite{deep-shallow-deep-crowdnet-16,deep-decide-18,deep-training-ICCNN-18,deep-switch-17,MCNN-16,SPooling-18,iccv-19}.
As illustrated in Figure \ref{fig:Multi-column}, the motivation of multi-column networks is intuitive.
Each column is devised with different receptive fields (e.g.,~different filter sizes) so that the features learned by different columns are expected to focus on different scales and resolutions.
By assembling features from all columns, multi-column networks are easily adaptive to the large variations of the scale due to the generalization ability across scales and resolutions.

\czq{Although multi-column architecture is naturally employed for addressing the issue of various scale change,}
previous works~\cite{CSRNet-18,deep-training-ICCNN-18,deep-pooling-fscale-Defense-18,SPooling-18,iccv-19} have pointed out that different columns always generate features with almost the same scale, which indicates that existing multi-column architectures cannot \ljx{effectively} improve the scale invariance of feature learning.
To further verify this observation, we have extensively analyzed three state-of-the-art networks, i.e.,~MCNN~\cite{MCNN-16}, CSRNet~\cite{CSRNet-18} and ic-CNN~\cite{deep-training-ICCNN-18}.
It is worth noting that CSRNet is a single column network, which has four different configurations~(i.e., different dilation rates).
We remould CSRNet to treat each configuration as a column, and design a four-column network as an alternative.
The Maximal Information Coefficient (MIC)\footnote{https://en.wikipedia.org/wiki/Maximal\_information\_coefficient} and the Structural SIMilarity (SSIM)\footnote{{https://en.wikipedia.org/wiki/Structural\_similarity}} are computed based on the results of different columns.
MIC measures the strength of association between the outputs (i.e.,~crowd counts)
and SSIM measures the similarity between density maps.
As shown in Table~\ref{tab:Multi-columns-problem}, 
\ljx{different }columns (Col.$\leftrightarrow$Col.) always output almost the same counts (i.e., high MIC) and the similar estimated density maps (i.e., high SSIM).
In contrast, a large gap between the ensemble of all columns and the ground truth (Col.$\leftrightarrow$GT.) still exists.
This comparison shows that there are substantial redundant parameters among columns, which makes multi-column architecture fails to learn the features across different scales.
On the other hand, it indicates that existing multi-column networks tend to overfit the data and can not learn the essence of the ground truth. 

Inspired by previous works~\cite{deep-training-ICCNN-18,CSRNet-18,deep-pooling-fscale-Defense-18}, we reveal that the problem of existing multi-column networks lies in the difficulty of learning features with different scales.
Generally speaking, there are two main problems:
1)~There is no supervision to guide multiple columns to learn features at different scales.
The current learning objective is only to minimize the errors of crowd count. 
Although we have designed different columns \czq{to} have different receptive fields, they are still gradually forced to generate features with almost the same scale along with the network optimization.
\ljx{2)~There are huge redundant parameters among columns.}
Because of parallel column architectures, multi-column networks naturally brought in redundant parameters.
As the analysis of~\cite{over-fitting-14}, with the increase of parameters, a more substantial amount of training data is also required.
It implies that existing multi-column networks are typically harder to train and easier to overfit.

\begin{table}[!t]
\small
\caption{\small The result analysis of three multi-column networks. The
values in the table are the average of all columns. Col.$\leftrightarrow$Col. is the
result between different columns. Col.$\leftrightarrow$GT is the result between
the ensemble of all columns and the ground truth.
}
\renewcommand\arraystretch{0.89}
\begin{center}
\vspace{-4mm}
\begin{tabular}{|l|c|c|c|c|} 
\hline
& \multicolumn{2}{c|}{\textbf{Col.$\leftrightarrow$ Col. }}      
& \multicolumn{2}{c|}{\textbf{Col.$\leftrightarrow$ GT }}  \\ 
\hline
\textbf{Method} & \textbf{MIC} & \textbf{SSIM} & \textbf{MIC} & \textbf{SSIM}\\ 
\hline \hline
\multicolumn{5}{|c|}{{ShanghaiTech Part A}~\cite{MCNN-16}}             \\ 
\hline
MCNN~\cite{MCNN-16}     &  0.94   & 0.71    &   0.52  &     0.55  \\
CSRNet~\cite{CSRNet-18}    &  0.93    &  0.84    &   0.74   &   0.71 \\
ic-CNN~\cite{deep-training-ICCNN-18}    &  0.92  & 0.72 & 0.70  & 0.68  \\ 
\hline\hline
\multicolumn{5}{|c|}{{UCF\_CC\_50}~\cite{r-mrf-count-multi-source-13}}             \\ 
\hline
MCNN~\cite{MCNN-16}       & 0.81  & 0.53  &  0.70 &   0.36    \\
CSRNet~\cite{CSRNet-18}   & 0.87  & 0.72  &  0.71  &  0.48 \\
ic-CNN~\cite{deep-training-ICCNN-18}    & 0.93  & 0.70  &  0.57  & 0.52      \\
\hline
\end{tabular}
\end{center}
\vspace{-6mm}
\label{tab:Multi-columns-problem}
\end{table}

In this paper, we propose a novel Multi-column Mutual Learning (McML) strategy to improve the learning of multi-column networks.
As illustrated in Figure~\ref{fig:framework}, our McML addresses the above two issues from two aspects.
1)~A statistical network is proposed to measure the mutual information between different columns.
The mutual information \czq{can} approximately \czq{measure} the scale correlation between features from different columns.
By additionally minimizing the mutual information in the loss, different column
\czq{structures} are forced to learn feature representations with different scales.
2)~Instead of the conventional optimization that updates the parameters of multiple columns simultaneously, we devise a mutual learning scheme that can alternately optimize each column while keeping the other columns fixed on each mini-batch training data. With such asynchronous learning \czq{steps}, each column is inclined to learn different feature representation from others, which can efficiently reduce the parameter redundancy and improve the generalization ability.
The proposed McML can be applied to all existing multi-column networks and is end-to-end trainable.
We conduct extensive experiments on four datasets to verify the effectiveness of our method.

The main contribution of this work is the proposal of Multi-column Mutual Learning (McML) strategy to improve the learning of multi-column networks.
The solution also provides the elegant views of how to explicitly supervise multi-column architectures to learn \ljx{features with different scales} and how to reduce the enormous redundant parameters and avoid overfitting, which are problems not yet fully understood in the literature.

\section{Related Work}
\label{sec:related-work}
\subsection{Detection-based Methods}
These models use visual object detectors to locate people in images. 
Given the individual localization of each people, crowd counting becomes trivial. 
There are two directions in this line, i.e., detection on 1) whole pedestrians \cite{d-bayesian-full-06,d-histograms-svm-full-05,d-haar-adaboost-full-01,d-full-bayesian-08} and 2) parts of pedestrians \cite{d-part2-08,d-part-15,d-harr-svm-part-01,d-part-11}. 
Typically, local features \cite{d-histograms-svm-full-05,d-harr-svm-part-01} are first extracted and then are exploited to train various detectors (e.g., SVM~\cite{d-harr-svm-part-01} and AdaBoost~\cite{d-adaboost-full-05}).
Although these works achieve satisfactory results for the low-density scenario, they are unable to generalize for high-density images since it is impossible to train a detector for extremely crowded scenes.

\vspace{-2mm}
\subsection{Regression-based Methods}
Different from detection-based models, regression-based methods directly estimate crowd count using image features.
It has two steps:~1)~extract powerful image features,~2)~use various regression models to estimate the crowd count.
Specifically, image features include edge features~\cite{r-edge-count-bayesian-12,r-edege-text-structure-count-12,r-count-edeger-crossing-Line-13,r-edge-count-96,r-density-edge-09} and texture features~\cite{r-count-13,r-edege-text-structure-count-12,r-mrf-count-multi-source-13,r-forest-density-15}. 
Regression methods cover Bayesian~\cite{r-edge-count-bayesian-12}, Ridge~\cite{r-edege-text-structure-count-12}, Forest~\cite{r-forest-density-15} and Markov Random Field~\cite{r-mrf-count-multi-source-13,r-mrf-count-01}.
Since these works always use handcrafted low-level features, they still cannot obtain satisfactory performance.

\vspace{-2mm}
\subsection{CNN-based Methods}
Due to substantial variations in the scale of people/head size, most recent studies extensively use Convolutional Neural Networks (CNN) with multi-column structures for crowd counting.
Specifically, a dual-column network is proposed by~\cite{deep-shallow-deep-crowdnet-16} to merge shallow and deep layers to estimate crowd counts. 
Inspired by this work, a great three-column network named MCNN is proposed by~\cite{MCNN-16}, which employs different filters on separate columns to obtain the various scale features. 
Noted that there are a lot of works to continually improve MCNN~\cite{deep-multi-attention-AFP-18,deep-fscale-multi-task-17,deep-multi-pyramid-cnn-17,deep-multi-cnn-boosting-16}.
Sam et al.~\cite{deep-switch-17} introduce a switching structure, which uses a classifier to assign input image patches to best column structures.
Recently, Liu et al.~\cite{deep-decide-18} propose a \ljx{m}ulti-column network to simultaneously estimate crowd density by detection and regression models.
Ranjan et al.~\cite{deep-training-ICCNN-18} employ a two-column structure to iterative train their model with different resolution images.

In addition to multi-column networks, there are a lot of methods to improve scale invariance of feature learning by
1)~studying on the fusion of multi-scale features~\cite{deep-fscale-Context-18,deep-fscal-pooling-PaDNet-18,deep-pooling-fscale-Defense-18,deep-multi-fusion-Adaptive-18},
2)~studying on multi-blob based scale aggregation networks~\cite{SANet-18,deep-multi-block-mscnn-17},
3)~designing scale-invariant convolutional or pooling layers~\cite{SPooling-18,CSRNet-18,deep-attention-cnn-ADCrowdNet-18,deep-multi-pyramid-cnn-17,deep-pooling-fscale-Defense-18}, 
and 4)~studying on automated scale adaptive networks~\cite{deep-auto-TDF-CNN-18,deep-auto-IGCNN-18,deep-fscale-multitask-SaCNN-18}. 
On the other hand, a lot of studies devote to using perspective maps~\cite{Perspective-18}, geometric constraints~\cite{deep-Geometric-18,deep-geo-Att-head-18}, and region-of-interest~\cite{deep-attention-cnn-ADCrowdNet-18} to further improve the counting accuracy.

These state-of-the-art methods aim to improve the scale invariance of feature learning.
Inspired by recent studies~\cite{deep-training-ICCNN-18,CSRNet-18,deep-pooling-fscale-Defense-18}, we reveal that existing multi-column networks cannot \ljx{effectively} learn different scale features as \Sec~\ref{sec:introduction}.
To solve this problem, we propose a novel Multi-column Mutual Learning (McML) strategy, which can be applied to all existing CNN-based multi-column networks and is end-to-end trainable.
It is noted that the previous work ic-CNN~\cite{deep-training-ICCNN-18} also proposes an iterative learning strategy to improve the learning of multi-column networks.
Different from our McML, since ic-CNN is designed for a specific neural architecture, it can not be generalized to all multi-column networks.
Additionally, we have tested our McML on the same network of ic-CNN.
Experimental results show that McML can still significantly improve the performance of the original ic-CNN.

\begin{figure}[!t]
\small
\begin{center}
\includegraphics[width=0.74\linewidth]{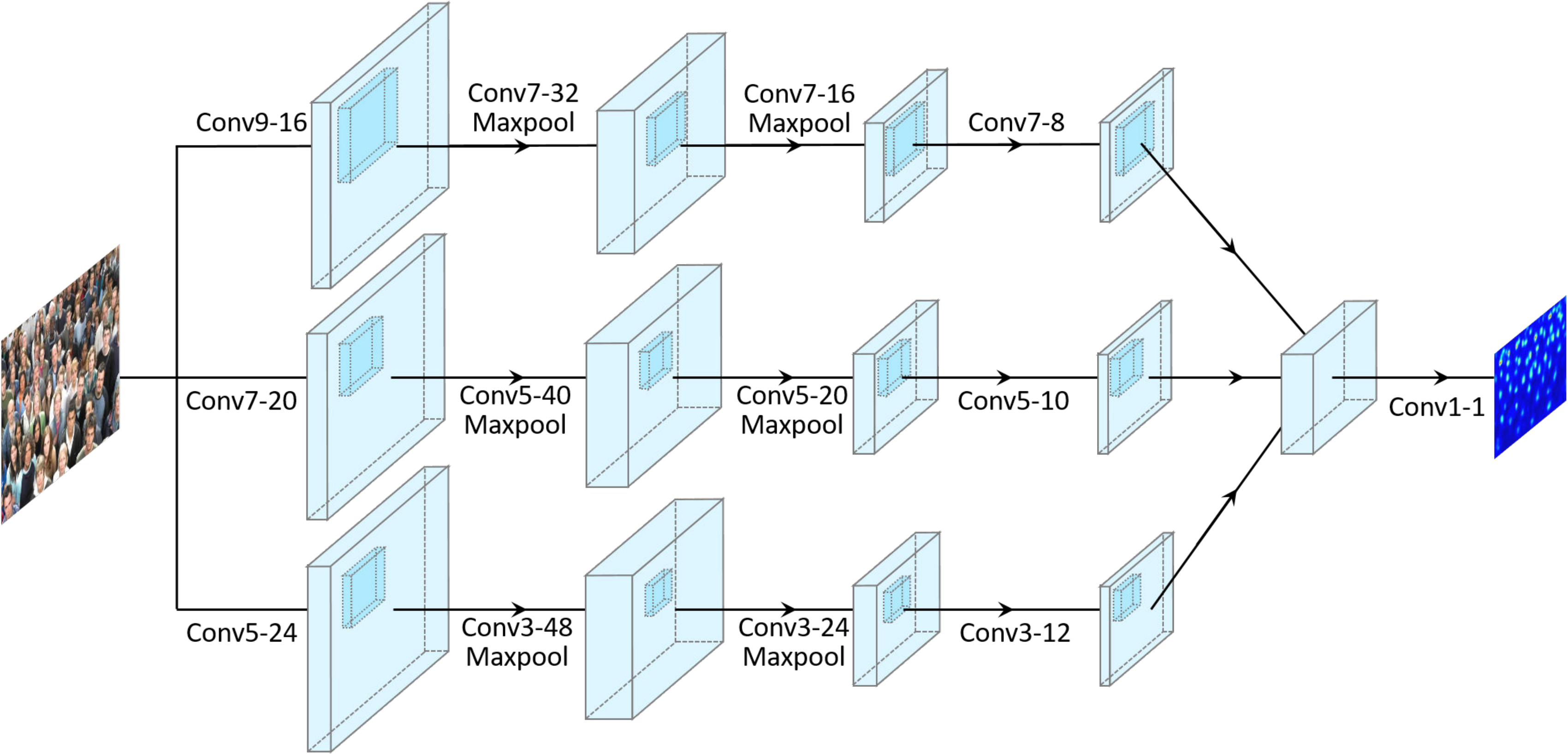}
\end{center}
\vspace{-4mm}
\caption{\small The architecture of MCNN~\cite{MCNN-16}. It is a classical Multi-column Convolutional Neural Network. It employs different \czq{size} of filters on three columns to obtain different scale features.}
\label{fig:Multi-column}
\vspace{-4mm}
\end{figure}

\section{Multi-column Mutual Learning}
\label{sec:our-method}
\czq{
In this section, we present the proposed Multi-column Mutual Learning (McML) strategy.
The problem formulation is first introduced in \Sec~\ref{sec:background-and-challenge}.
Then the overview of our McML is described in \Sec~\ref{sec:overview-of-McML}.
More details of McML are illustrated in \Sec~\ref{sec:mutual-information-estimation} to \ref{sec:network-architecture}.}

\subsection{Problem Formulation}
\label{sec:background-and-challenge}
Recent studies define crowd counting task is as a density regression problem~\cite{SANet-18,nips-10,MCNN-16}.
Given $N$ training images $\textbf{X}=\{x_{1}, \cdots, x_{N}\}$ as the training set, each image $x_{i}$ is annotated with a total of $c_{i}$ center points of pedestrians' heads $\textbf{P}_{i}^{gt}=\{P_{1}, P_{2},\cdots, P_{c_{i}}\}$.
Typically, the ground truth density map $y_{i}$ of image $x_{i}$ is generated as, 
\beq
\label{eq:ground-truth-density}
\vspace{-1mm}
\forall p\in x_{i}, y_{i}=\sum_{P\in\textbf{P}_{i}^{gt}}\NN^{gt}(p;\mu=P,\sigma^{2}),
\eeq
where $p$ is a pixel and $\NN^{gt}$ is a Gaussian kernel with standard deviation $\sigma$. 
The number of people $c_{i}$ in image $x_{i}$ is equal to the sum of density of all pixels as $\sum_{p\in x_{i}}y_{ i}(p)=c_{i}$. 
With these training data, crowd counting models aim to learn a regression model \ljx{$G$} with parameters $\theta$ to minimize the difference between estimated density map \ljx{$G_{\theta}(x_i)$} and ground truth density map $Y_i$. 
Specifically, Euclidean distance, i.e.,~$L_2$ loss is employed to get an approximate solution,
\beq
\label{eq:l-2-2}
\vspace{-1mm}
L_2= \frac{1}{2N} \sum^{N}_{i=1} (\ljx{G}_{\theta}(x_i)-y_i)^2,
\eeq
\czq{where as the size of input images are different, the value of \Eq~\ref{eq:l-2-2} is further normalized by the number of pixels in each image.}

\czq{It is noted that, as shown in \Fig~\ref{fig:scale-changes}, enormous variation in the scale of people/head size is a critical problem for crowd counting.
Many studies~\cite{single-error-05,single-error-14,single-error-16,MM-2018,single-error-17,huang2018gnas} have proved that only using an individual regression model is theoretically far from the global optimal.
To improve the scale invariance of feature learning, Convolutional Neural Networks with multi-column structures are extensively studied by recent works~\cite{deep-multi-attention-AFP-18,deep-fscale-multi-task-17,deep-multi-pyramid-cnn-17,deep-multi-cnn-boosting-16,MCNN-16}.
\Fig~\ref{fig:Multi-column} illustrates a typical multi-column network named MCNN~\cite{MCNN-16}.
The intentions of multi-column networks are natural, where each column structure is devised with different receptive fields (e.g.,~different filter sizes) so that the features learned by individual column is expected to focus on a particular scale of people/head size.
With the ensemble of features from all columns, multi-column networks are easily adaptive to handle the large scale variations.}

\czq{Although the motivation for multi-column structures is straightforward, previous works~\cite{CSRNet-18,deep-training-ICCNN-18,deep-pooling-fscale-Defense-18} have pointed out that existing multi-column networks cannot improve the scale invariance of features learning.
As analyzed in \Sec~\ref{sec:introduction}, we are convinced that there are abundant redundant parameters between columns, which causes multi-column structures to fail to learn the features across different scales and invariably get almost the same estimated crowd counts and density maps.
After thoroughly surveying previous works~\cite{CSRNet-18,deep-training-ICCNN-18,deep-pooling-fscale-Defense-18,single-error-16,single-error-17} and analyzing our experimental results in \Tbl~\ref{tab:Multi-columns-problem}, we further reveal that the main problem of existing multi-column networks lies in the learning process.
Generally speaking, current learning strategy has two main weaknesses.
1)~It only optimizes the objective of crowd counting, while completely ignores the intention of using multi-column structures to learn different scale features.
2)~It instantly optimizes multi-column structures at the same time, which can result in the enormous redundant parameters among columns and overfitting on the limited training data.
To address these problems, our work aims to propose a general learning strategy \ljx{named} Multi-column Mutual learning (McML) to improve the learning of multi-column networks.}

\begin{figure*}
\small
\begin{center}
\includegraphics[width=0.8\textwidth,height=0.27\textheight]{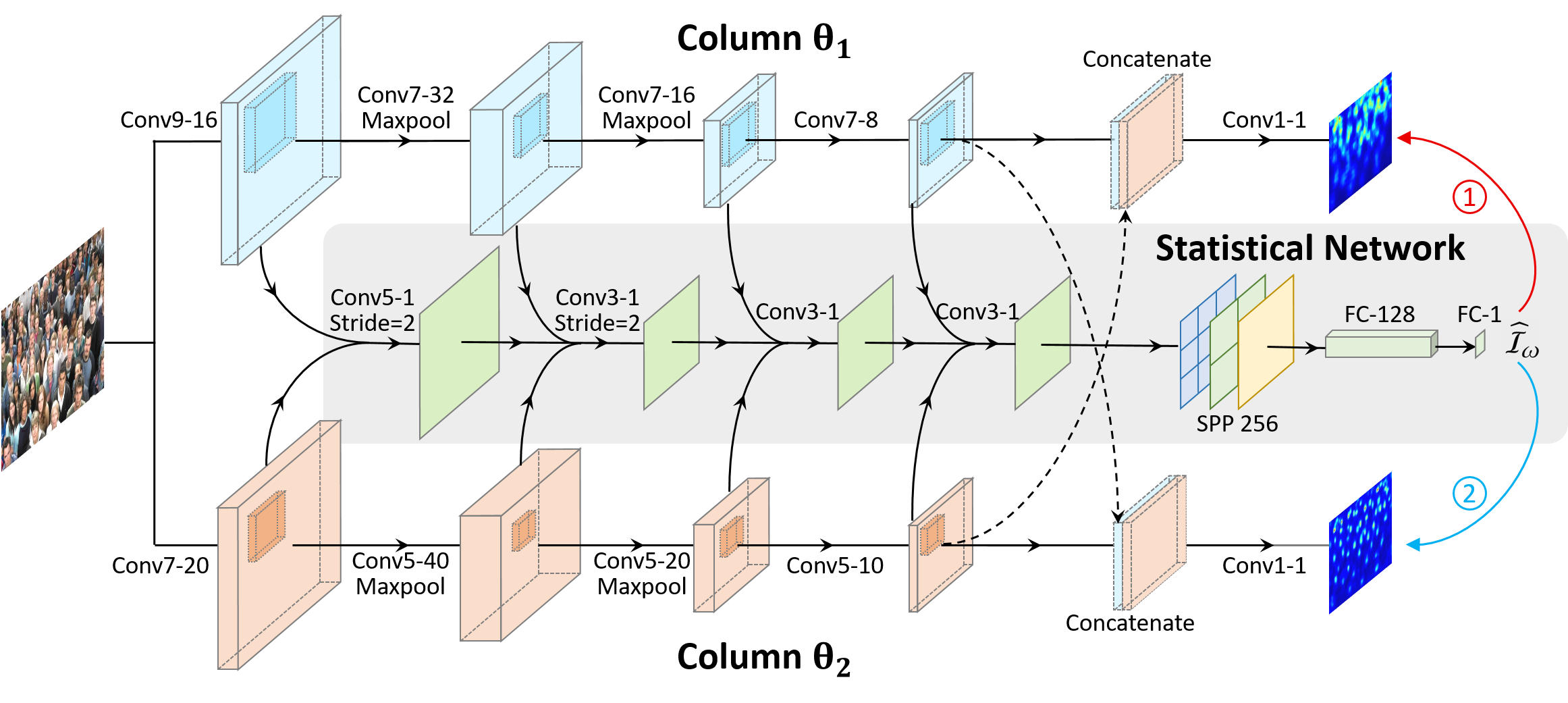}
\end{center}
\vspace{-5mm}
\caption{\small Overview of our Multi-column Mutual Learning (McML) strategy. 
It is equivalent to adding a statistical network to estimate the mutual information $\widehat{\II}_{\omega}$ between columns.
By minimizing the mutual information, it can guide \ljx{m}ulti-columns to learn different scale information.
Additionally, McML is a mutual learning \ljx{scheme} as arrows \ding{172} and \ding{173},
\czq{where each column is alternately optimized while keeping the other columns fixed on each mini-batch training data.}
Specifically, this is an example of two columns ($\rm \Theta_{1}$ and $\rm \Theta_{2}$) in MCNN~\cite{MCNN-16}.
ConvX-Y implies a convolution layer has Y filters with X$\times$X kernel size.
The stride of all convolutional layers is 1, except for the special reminder.
MaxPool is the max pooling layer with a stride of 2. \ljx{[Best viewed in color].}}
\label{fig:framework}
\vspace{-4mm}
\end{figure*}

\vspace{-2mm}
\subsection{Overview of McML}
\label{sec:overview-of-McML}
\czq{In this section, we present an overview of Multi-column Mutual Learning (McML) strategy.
For the sake of simplicity, we introduce the case of two columns as an example.
As shown in \Fig~\ref{fig:framework}, our McML has two main innovations.
\begin{itemize}
\item McML has integrated a statistical network into multi-column structures to automatically estimate the mutual information between columns.
The essential of the statistical network is a classifier network.
Specifically, the input\ljx{s} are features from different columns, and the output is the mutual information between columns.
We use mutual information to approximately indicate the scale correlation between features from different columns.
By minimizing the mutual information between columns, McML can guide each column to focus on different image scale information.
\vspace{1mm}
\item McML is a mutual learning \ljx{scheme}.
Different from updating the parameters of multiple columns simultaneously, McML alternately optimizes each column in turn until the network converged.
In the learning of each column, the mutual information between columns is first estimated as prior knowledge to guide the parameter update.
With the help of the mutual information between columns, McML can alternately make each column to be guided by other columns to learn different image scales/resolutions.
It is proved that this mutual learning \ljx{scheme} can significantly reduce the volume of redundant parameters and avoid overfitting.
\end{itemize}}

\czq{Mathematically, two columns with parameters ${\theta_1}$ and ${\theta_2}$ are alternately trained as,}
\beq
\label{eq:column-1}
L_{\theta_1}=\min_{\theta_1} L_2(\mbox{Conv}(F_{\theta_1 \circ \theta_2}(X)),Y)+\alpha \widehat{\II}_{\omega}(C_{\theta_1}; C_{\theta_2}),
\eeq
\beq
\label{eq:column-2}
L_{\theta_2}=\min_{\theta_2} L_2(\mbox{Conv}(F_{\theta_1 \circ \theta_2}(X)),Y)+\alpha \widehat{\II}_{\omega}(C_{\theta_1}; C_{\theta_2}),
\eeq
\czq{where each column is trained by two losses.
$L_2$ loss (\Eq~\ref{eq:l-2-2}) is used to minimize counting errors, and $\widehat{\II}_{\omega}$ (\Eq~\ref{eq:DV-sample}) is employed to minimize the mutual information between columns.
$\alpha$ is the weight to trade off two losses.
The value of mutual information $\widehat{\II}_{\omega}$ is computed by the statistical network with parameters $\omega$.
Here we have slightly abused symbols.
$C_{\theta_1}$ and $C_{\theta_2}$ are features from different convolutional layers of two columns, which are used to estimate the mutual information.
$F_{\theta_1 \circ \theta_2}(X)$ means the ensemble (i.e.,~\czq{concatenation}) of features at the last convolutional layers for two both columns.
Conv is a $1 \times 1$ convolutional layer that is used to predict density maps for crowd counting.}

\czq{Typically, our proposed McML is also a mutual learning \ljx{scheme}.
Two columns are alternately optimized until convergence.
In the learning of each column, the mutual information $\widehat{\II}_{\omega}$ is first estimated as prior knowledge to guide the parameter update.
Once the optimization of one column (e.g.,~$\theta_1$) is finished, we will update the mutual information $\widehat{\II}_{\omega}$ again and alternately to update the other column (e.g.,~$\theta_2$).
Additionally, it is noted that \Eqs~\ref{eq:column-1} and \ref{eq:column-2} show the situation of most multi-column networks (e.g.,~CrowdNet~\cite{deep-shallow-deep-crowdnet-16}, AM-CNN~\cite{deep-geo-Att-head-18}, and MCNN~\cite{MCNN-16}), where the features of multi-columns are concatenated to estimate density maps.
However, a few multi-column networks (e.g.,~ic-CNN~\cite{deep-training-ICCNN-18}) predict density maps in all columns.
In these cases, $F_{\theta_1 \circ \theta_2}$ of \Eqs~\ref{eq:column-1} and \ref{eq:column-2} should be replaced with $F_{\theta_1}$ and $F_{\theta_2}$ respectively.
Where $F_{\theta_1}$ and $F_{\theta_2}$ are features from the last convolutional layers at two columns.}

Specifically, we will introduce the mutual information estimation (i.e., computation of $\widehat{\II}_{\omega}$) in \Sec~\ref{sec:mutual-information-estimation}, the mutual learning \ljx{scheme} in \Sec~\ref{sec:mutual-learning-process} and neural architectures of statistical networks in \Sec~\ref{sec:network-architecture}.

\subsection{Mutual Information Estimation}
\label{sec:mutual-information-estimation}
\czq{In this section, we first briefly introduce the definition of mutual information.
Then we present the statistical network in details.}

\czq{Mutual information is a fundamental quantity for measuring the correlation between variables.
We treat column structures as different variables.
Inspired by the success of previous works~\cite{mr1,mr2}, we use mutual information to indicate the degree of parameter redundancy between columns.
Moreover, mutual information can also approximately measure the scale correlation between features from different columns.
Instead of estimating the mutual information with parameters of columns, similar to~\cite{mr4,ICMR-17,MI-use-18}, we chooses to compute the mutual information using the features of multi-columns since our objective is to learn different scale features.
Typically, the mutual information between features $C_{\theta_1}$ and $C_{\theta_2}$ is defined as,}
\beq
\label{eq:im-definition}
\II(C_{\theta_1};C_{\theta_2}) := H(C_{\theta_1}) - H(C_{\theta_1} \mid C_{\theta_2}),
\eeq
where $H$ is the Shannon entropy.
$H(C_{\theta_1}\, |\, C_{\theta_2})$ measures the uncertainty in $C_{\theta_1}$ given $C_ {\theta_2}$.
Previous works \cite{mr1,mr2,mr3,mr4} widely use Kullback-Leibler (KL) divergence to compute the mutual information,
\beq 
\label{eq:kl-mi}
\II(C_{\theta_1};C_{\theta_2}) = \KL{\PJ{C_{\theta_1}}{C_{\theta_2}}}{\PI{C_{\theta_1}}{C_{\theta_2}}},
\eeq
\czq{where $\PJ{C_{\theta_1}}{C_{\theta_2}}$ is the joint distribution of two features.
$\PP_{ C_{\theta_1}}$ and $\PP_{ C_{\theta_2}}$ are the marginal distributions.
$\otimes$ means the production.
Since the joint distribution $\PJ{C_{\theta_1}}{C_{\theta_2}}$ and the product of marginal distributions $\PI{C_{\theta_1}}{C_{\theta_2}}$ are unknown in our case, the mutual information of two columns is challenging to compute~\cite{mutual-information-difficult-03}.}

Fortunately, inspired by the previous work named MINE~\cite{Mime}, we propose a statistical network to estimate the mutual information.
The essence of the statistical network is a classifier.
\czq{It can be used to distinguish the samples between the joint distribution and the product of marginal distributions.}
Instead of computing \Eq~\ref{eq:kl-mi}, the statistical network \czq{chooses} to use Donsker-Varadhan representation~\cite{DV-83} i.e.,~$\II(C_{\theta_1}; C_{\theta_2}) \geq \widehat{\II}_{\omega}(C_{\theta_1}; C_{\theta_2})$, to get a lower-bound \czq{for the mutual information estimation},
\beq 
\label{eq:DV-sample}
\widehat{\II}_\omega \gets \frac{1}{b} \sum_{i=1}^{b} T_\omega (C_{\theta_1}^{(i)},C_{\theta_2}^{(i)}) - \log(\frac{1}{b} \sum_{i=1}^{b} e^{T_\omega(C_{\theta_1}^{(i)},\widehat{C_{\theta_2}^{(i)})}}),
\eeq
\czq{where $T_\omega$ is the statistical network with parameters $\omega$.
To compute the lower-bound $\widehat{\II}\ljx{_\omega}$, we randomly select $b$ training images. 
With the forward pass of the network, we directly get $b$ pairs of features from two column structures as the joint distribution $\small{(C_{\theta_1}, C_{\theta_2}) \sim \PJ{C_{\theta_1}}{C_{\theta_2}}}$.
At the same time, we randomly disrupt the order of $C_{\theta_2}$ in $\small{(C_{\theta_1}, C_{\theta_2}) \sim \PJ{C_{\theta_1}}{C_{\theta_2}}}$
to get $b$ pairs of features as the product of the marginal distribution $\small{(C_{\theta_1}, \widehat{C_{\theta_2}})\sim \PI{C_{\theta_1}}{C_{\theta_2}}}$.
Then we input these features to the statistical network $T_\omega$.
By calculating the $b$ outputs of the statistical network as \Eq~\ref{eq:DV-sample}, we can get a lower-bound for the mutual information estimation.
Here we use moving average to get the gradient of \Eq~\ref{eq:DV-sample}.
By maximizing this lower-bound, we can approximately obtain the real mutual information.
More details of the mutual information estimation are provided in \Alg~\ref{alg:mi_donsker_estimation}.}

\czq{Without loss of generality, the statistical network $T_\omega$ can be designed as any classifier networks according to the different multi-column networks.
We have tested McML on three multi-column networks (i.e.,~MCNN~\cite{MCNN-16}, CSRNet~\cite{CSRNet-18} and ic-CNN~\cite{deep-training-ICCNN-18}). 
The statistical networks for these baselines are described in \Sec~\ref{sec:network-architecture}.}

\setlength{\textfloatsep}{9pt}
\begin{algorithm}[!t]
\small
\begin{algorithmic}[1]
\INPUT Randomly sampled $b$ images.
\STATE Draw features from two columns as the joint distribution, 
\STATE $(C_{\theta_1}^{(1)}, C_{\theta_2}^{(1)}), \ldots, (C_{\theta_1}^{(b)}, C_{\theta_2}^{(b)}) \sim \PJ{C_{\theta_1}}{C_{\theta_2}}$;
\STATE \czq{Randomly disrupt $C_{\theta_2}$ as the product of marginal distribution,}
\STATE $(C_{\theta_1}^{(1)},\widehat{C_{\theta_2}^{(1)}}), \ldots, (C_{\theta_1}^{(b)},\widehat{C_{\theta_2}^{(b)}}) \sim \PI{C_{\theta_1}}{C_{\theta_2}}$;
\STATE Evaluate mutual information $\widehat{\II}_\omega$ Das \Eq~\ref{eq:DV-sample};
\STATE Use moving average to get the gradient,
\STATE $\widehat{G}(\omega) \gets \widetilde{\nabla}_{\omega} \widehat{\II}_\omega$;
\STATE Update the statistic\ljx{al} network parameters,
\STATE $\omega \gets \omega + \widehat{G}(\omega)$; 
\end{algorithmic} 
\caption{Mutual Information Estimation 
\label{alg:mi_donsker_estimation}}
\end{algorithm}

\subsection{Mutual Learning Scheme}
\label{sec:mutual-learning-process}
\czq{Our proposed McML is a mutual learning \ljx{scheme}.
For the sake of simplicity, we present the case of two columns as an example.
As shown in \Alg~\ref{alg:mutual-learning}, we alternately optimize two columns in each mini-batch until convergence.
In each learning iteration, we randomly sample $b$ training images.
Before optimizing column $\theta_1$, the mutual information is first estimated as prior knowledge to guide the parameter update.
With forward of the network, the features of two column structures are sampled to update the statistical network $T_{\omega}$ and estimate the mutual information $\widehat{\II}_{\omega}$ as \Alg~\ref{alg:mi_donsker_estimation}.
With the guidance of the mutual information, our McML can supervise column $\theta_1$ to learn as much as possible different scale features from column $\theta_2$.
It is noted that we have fixed parameters of other columns (i.e.,~$\theta_2$) and statistic\ljx{al} network ($T_\omega$), and only update column $\theta_1$.
Since the size of input images are different, we have to update column structure on each image.
After back-propagation of a total of $b$ images, the column $\theta_2$ will be optimized in similar steps.}

It is noted that our McML can be naturally extended to multi-columns architectures. 
For the case of $K>2$, the loss function of a column $\theta_k$ is computed as,
\vspace{-2mm}
\beq
\normalsize
\label{eqn:mutual-column-k}
L_{\theta_{k}} = L_2(\mbox{Conv}(F_{\theta_1 \circ \theta_2,..,\theta_K}(X)),Y)+\frac{\alpha}{K-1}\sum_{l=1, k \neq l}^{K} \widehat{\II}_{\omega}(C_{\theta_l}; C_{\theta_k}).
\eeq
Similar to \Eqs~\ref{eq:column-1} and \ref{eq:column-2}, where $F_{\theta_1 \circ \theta_2,..,\theta_K}$ means the ensemble (i.e., concanation) of features from the last convolutional layers at \ljx{m}ulti-columns. 
$C_{\theta_{*}}$ is the features from different convolutional layers at each column.
$\alpha$ is a weight to trade off two losses.
At this point, we only need to add more steps to estimate mutual information of multi-columns.
Once the mutual information is obtained, multi-column structures are still alternately optimized until convergence.

\begin{table}[!b]
\small
\caption{\small The structure of statistic\ljx{al} network.
The convolutional, spatial pyramid pooling, and fully connected layers are denoted as "Conv (kernel size)-(\ljx{number of channels})-(stride)", "SPP (\ljx{size of outputs})", and "FC (\ljx{size of outputs})".
}
\renewcommand\arraystretch{0.9}
\begin{center}
\vspace{-2mm}
\czq{\begin{tabular}{|l|l|l|} 
\hline
\multicolumn{1}{|c|}{\textbf{MCNN}~\cite{MCNN-16}} & \multicolumn{1}{c|}{\textbf{CSRNet}~\cite{CSRNet-18}} & \multicolumn{1}{c|}{\textbf{ic-CNN}~\cite{deep-training-ICCNN-18}}  \\ 
\hline \hline
Conv 5-1-2    & Conv 3-1-1     & Conv 3-1-1      \\
Conv 3-1-2    & Conv 3-1-1     & Conv 3-1-1      \\
Conv 3-1-1    & Conv 3-1-1     & Conv 3-1-1      \\
Conv 3-1-1    & Conv 3-1-1     & Conv 3-1-1      \\
SPP 256       & Conv 3-1-1     & Conv 3-1-1     \\
FC 128      & Conv 3-1-1     & SPP 256         \\
FC 1        & SPP 256        & FC 128      \\
              & FC 128       & FC 1        \\
              & FC 1         &                  \\
\hline
\end{tabular}}
\end{center}
\vspace{-2mm}
\label{tab:statistics-network}
\end{table}

\setlength{\textfloatsep}{9pt}
\begin{algorithm}[!t]
\small
\begin{algorithmic}[1]
\INPUT Training set $X$, Ground truth $Y$.
\STATE $\theta_1$, $\theta_2$ and $\omega \gets \text{initialize network parameters}$;
\REPEAT
\STATE Randomly sampled $b$ images from $X$;
\czq{\STATE Estimate mutual information $\widehat{\II}_{\omega}$ and update statistical network $T_{\omega}$ as \Alg~\ref{alg:mi_donsker_estimation};}
\STATE \czq{Update column $\theta_{1}$ as~\Eq~\ref{eq:column-1} on each image,} 
\STATE $\theta_{1} \leftarrow \theta_{1} +  \frac{\partial L_{{\theta_{1}}}}{\partial \theta_{1}}$;
\czq{\STATE Estimate mutual information $\widehat{\II}_{\omega}$ and update statistical network $T_{\omega}$ as \Alg~\ref{alg:mi_donsker_estimation};}
\STATE \czq{Update column $\theta_{2}$ as~\Eq~\ref{eq:column-2} on each image,} 
\STATE $\theta_{2} \leftarrow \theta_{2} +  \frac{\partial L_{{\theta_{2}}}}{\partial \theta_{2}}$;
\UNTIL{Convergence}
\end{algorithmic} 
\caption{Mutual Learning Scheme}
\label{alg:mutual-learning}
\end{algorithm}

\vspace{-2mm}
\subsection{Network Architectures}
\label{sec:network-architecture}
We employ McML to improve three state-of-the-art networks, including MCNN~\cite{MCNN-16}, CSRNet~\cite{CSRNet-18}, and ic-CNN~\cite{deep-training-ICCNN-18}.
Table~\ref{tab:statistics-network} shows the neural architecture of statistical networks.
To better understand the details, Figure~\ref{fig:framework} gives a real example of two columns in MCNN~\cite{MCNN-16}.
With sharing the parameters, no matter how many columns are adopted, each \ljx{m}ulti-column network only needs one single statistical network.
The inputs of statistical networks are the features from different layers.
We use convolutional layers with one output channel to reduce the feature dimension.
Since training images have different size and inspired by the previous work \cite{CVPR-2017}, one spatial pyramid pooling (SPP) layer is applied to reshape the features from the last convolutional layer into a fixed dimension.
Finally, two fully connected layers are employed as a classifier.
Similar to \cite{Mime}, Leaky-ReLU~\cite{lrelu} is used as the activation function for all convolutional layers, and no activation function for other layers.

\begin{table*}[t]
\small
\caption{Performance of ablation studies. Comparison of Org. (Original Baseline), MLS (Mutual Learning Scheme), MIE (Mutual Information Estimation), and McML (Multi-column Mutual Learning) on four crowd counting datasets.}
\renewcommand\arraystretch{0.95}
\begin{center}
\vspace{-3mm}
\begin{tabular}{|l|cc|cc|cc|cc|c|}
\hline
& \multicolumn{2}{c|}{\textbf{ShanghaiTech A~\cite{MCNN-16}}} & \multicolumn{2}{c|}{\textbf{ShanghaiTech B~\cite{MCNN-16}}} & \multicolumn{2}{c|}{\textbf{UCF\_CC\_50~\cite{r-mrf-count-multi-source-13}}} & \multicolumn{2}{c|}{\textbf{UCSD~\cite{deep-crowd-scene-15}}} & \textbf{WorldExpo'10~\cite{r-count-privacy-preserving-08}} \\ \hline
\textbf{Method} & \textbf{MAE}         & \textbf{MSE}          & \textbf{MAE}         & \textbf{MSE}          & \textbf{MAE}        & \textbf{MSE}        & \textbf{MAE}     & \textbf{MSE}    & \textbf{MAE}          \\ \hline \hline
MCNN~\cite{MCNN-16}            & 110.2                & 173.2                 & 26.4                 & 41.3                  & 377.6               & 509.1               & 1.07             & 1.35            & 11.6                  \\
MCNN+MLS        & 105.2                & 160.3                 & 22.2                 & 34.2                  & 332.8               & 425.3               & 1.04             & 1.35            & 10.8                  \\
MCNN+MIE        & 106.7                & 160.5                 & 25.4                 & 35.6                  & 338.6               & 447.4               & 1.12             & 1.47            & 11.0                  \\
MCNN+McML       & 101.5                & 157.7                 & 19.8                 & 33.9                  & 311.0               & 402.4               & 1.03             & 1.24            & 10.2                  \\ \hline\hline
CSRNet~\cite{CSRNet-18}          & 68.2                 & 115.0                 & 10.6                 & 16.0                  & 266.1               & 397.5               & 1.16             & 1.47            & 8.6                   \\
CSRNet+MLS      & 64.2                 & 109.3                 & 9.9                  & 12.3                  & 254.2               & 376.3               & \textbf{1.00}    & 1.31            & 8.4                   \\
CSRNet+MIE      & 65.6                 & 111.0                 & 9.3                  & 12.8                  & 264.9               & 387.1               & 1.06             & 1.40            & 8.3                   \\
CSRNet+McML     & \textbf{59.1}        & \textbf{104.3}        & \textbf{8.1}         & \textbf{10.6}         & 246.1               & 367.7               & 1.01             & 1.27            & \textbf{8.0}          \\ \hline\hline
ic-CNN~\cite{deep-training-ICCNN-18}          & 68.5                 & 116.2                 & 10.7                 & 16.0                  & 260.9               & 365.5               & 1.14             & 1.43            & 10.3                  \\
ic-CNN+MLS      & 67.4                 & 112.8                 & 10.3                  & 14.6                  & 248.4               & 364.3               & 1.02             & 1.28            & 9.7                   \\
ic-CNN+MIE      & 66.3                 & 111.8                 & 11.3                 & 15.1                  & 255.3               & 368.2               & 1.06             & 1.34            & 9.8                   \\ 
ic-CNN+McML     & 63.8                 & 110.5                 & 10.1                 & 13.9                  & \textbf{242.9}      & \textbf{357.0}      & \textbf{1.00}    & \textbf{1.20}   & 8.5                   \\ \hline
\end{tabular}
\end{center}
\vspace{-4mm}
\label{tab:ablation-study}
\end{table*}


Specifically, MCNN adopts 3 column \ljxa{structures}.  
Each column contains 4 convolutional layers.
Intuitively,  the statistical network of MCNN uses 4 convolutional layers to embed the features as Figure ~\ref{fig:framework}.
CSRNet is a single column network.
The first 10 convolutional layers are from pre-trained VGG-16~\cite{vgg-14}.
The last 6 dilated convolutional layers are utilized to estimate the crowd count\ljx{s}.
The original version has 4 configurations for 6 dilated convolutional layers~(i.e.,~different dilation rates).
Here we treat 4 configurations as 4 different columns.
Similarly, as shown in Table ~\ref{tab:statistics-network}, the statistical network of CSRNet utilizes 6 convolutional layers to embed the features for 6 dilated convolutional layers in each column.
ic-CNN contains two columns~(i.e.,~Low Resolution (LR) and High Resolution (HR) columns). 
LR contains 11 convolutional layers and 2 max-pooling layers, and HR has 10 convolutional layers with 2 max-pooling layers and 2 deconvolutional layers. 
As Table~\ref{tab:statistics-network} shows, the statistical network of ic-CNN uses \czq{5} convolutional layers to embed features from corresponding \czq{5} convolutional layers after the second max pooling layer at both columns.

\vspace{-2mm}
\section{Experiment}
\label{sec:experiment}
\subsection{Experiment Settings}
\label{sec:experiment-settings}
\noindent\textbf{Datasets}.~To evaluate the effectiveness of our McML, we conduct experiments on four crowd counting datasets, i.e.,
\noindent ShanghaiTech~\cite{MCNN-16},
UCF\_CC\_50~\cite{r-mrf-count-multi-source-13},
UCSD~\cite{r-count-privacy-preserving-08}, and
WorldExpo'10~\cite{deep-crowd-scene-15}.
Specifically, ShanghaiTech dataset consists of two parts: Part\_A and Part\_B. 
Part\_A is collected from the internet and usually has very high crowd density. 
Part\_B is from busy streets and has a relatively sparse crowd density.
UCF\_CC\_50 is mainly collected from Flickr and contains images of extremely dense crowds.
UCSD and WorldExpo'10 are both collected from actual surveillance cameras and have low resolution and sparse crowd density.
More details of datasets split are illustrated in supplementary material.

\noindent\textbf{Learning Settings}.~We use our McML to improve MCNN, CSRNet, and ic-CNN.
For MCNN, the network is initiated by a Gaussian distribution with a mean of 0 and a standard deviation of 0.01. 
Adam optimizer\ljx{~\cite{adam-14}} with a learning rate of 1e-5 is used to train three columns. 
For CSRNet, the first 10 convolutional layers are fine-tuned from the pre-trained VGG-16~\cite{vgg-14}. 
The other layers are initiated in the same way as MCNN. 
We use Stochastic gradient descent (SGD) with a fixed learning rate of 1e-6 to finetune four columns.
For ic-CNN, input features from Low-resolution column to High-resolution column are neglected.
The SGD with the learning rate of 1e-4 is used to train two columns. 
The learning settings of the statistical network for all baselines are the same.
The number of samples $b$ is 75.
Moving average is used to evaluate gradient bias.
Adam optimizer with a learning rate of $1e-4$ is used to optimize the statistic\ljx{al} network.
More details of ground truth generation and data augmentation are illustrated in supplement materials.

\noindent\textbf{Evaluation Details}.~
Following previous works~\cite{MCNN-16, CSRNet-18, deep-training-ICCNN-18},
we use mean absolute error~(MAE) and mean square error~(MSE) to evaluate the performance:
\vspace{-2mm}
\beq
\normalsize
\label{eq:mae-mse}
\ljx{MAE=\frac{1}{N}\sum_{i=1}^N\left | Z_i-Z_i^{gt} \right |,\;\;
MSE=\sqrt{\frac{1}{N}\sum_{i=1}^N\left ( Z_i-Z_i^{gt} \right )^2},}
\eeq
\ljx{where $Z_i$ is the estimated crowd count and $Z_i^{gt}$ is the ground truth count of the i-th image.}
$N$ is the number of test images. 
The MAE indicates the accuracy of the estimation, while the MSE indicates the robustness.

\begin{table*}[t]
\caption{\small Comparison with state-of-the-art methods on ShanghaiTech~\cite{MCNN-16}, UCF\_CC\_50~\cite{r-mrf-count-multi-source-13} and UCSD~\cite{deep-crowd-scene-15} datasets.}
\small
\renewcommand\arraystretch{0.95}
\begin{center}
\vspace{-2mm}
\begin{tabular}{|lll|cc|cc|cc|cc|}
\hline
&     &      & \multicolumn{2}{c|}{\textbf{ShanghaiTech A}~\cite{MCNN-16}} & \multicolumn{2}{c|}{\textbf{ShanghaiTech B}~\cite{MCNN-16}} & \multicolumn{2}{c|}{\textbf{UCF\_CC\_50}~\cite{r-mrf-count-multi-source-13}} & \multicolumn{2}{c|}{\textbf{UCSD}~\cite{deep-crowd-scene-15} }     \\ \hline
\multicolumn{1}{|l|}{\textbf{Method}} & \multicolumn{2}{l|}{\textbf{Venue \& Year}}     & \textbf{MAE}              & \textbf{MSE}             & \textbf{MAE}             & \textbf{MSE}              & \textbf{MAE}             & \textbf{MSE}            & \textbf{MAE}           & \textbf{MSE}           \\ \hline \hline
\multicolumn{1}{|l|}{Idrees et al.~\cite{r-mrf-count-multi-source-13}}   & CVPR                  & 2013 & -                & -               & -               & -                & 419.5           & 541.6          & -             & -             \\
\multicolumn{1}{|l|}{Zhang et al.~\cite{deep-crowd-scene-15}}    & CVPR                  & 2015 & 181.8            & 277.7           & 32.0            & 49.8             & 467.0           & 498.5          & 1.60          & 3.31          \\
\multicolumn{1}{|l|}{CCNN~\cite{deep-multi-perspetive-free-16}}            & ECCV                  & 2016 & -                & -               & -               & -                & -               & -              & 1.51          & -             \\
\multicolumn{1}{|l|}{Hydra-2s~\cite{deep-multi-perspetive-free-16}}        & ECCV                  & 2016 & -                & -               & -               & -                & 333.7           & 425.3          & -             & -             \\
\multicolumn{1}{|l|}{C-MTL~\cite{deep-fscale-multi-task-17}}           & AVSS                  & 2017 & 101.3            & 152.4           & 20.0            & 31.1             & 322.8           & 397.9          & -             & -             \\
\multicolumn{1}{|l|}{SwitchCNN~\cite{deep-switch-17}}       & CVPR                  & 2017 & 90.4             & 135.0           & 21.6            & 33.4             & 318.1           & 439.2          & 1.62          & 2.10          \\
\multicolumn{1}{|l|}{CP-CNN~\cite{deep-multi-pyramid-cnn-17}}          & ICCV                  & 2017 & 73.6             & 106.4           & 20.1            & 30.1             & 295.8           & \textbf{320.9}          & -             & -             \\
\multicolumn{1}{|l|}{Huang at al.~\cite{deep-body-BSAD-18}}    & TIP & 2018 & -                & -               & 20.2            & 35.6             & 409.5           & 563.7          & \textbf{1.00}         & 1.40          \\
\multicolumn{1}{|l|}{SaCNN~\cite{deep-fscale-multitask-SaCNN-18}}           & WACV                  & 2018 & 86.8             & 139.2           & 16.2            & 25.8             & 314.9           & 424.8          & -             & -             \\
\multicolumn{1}{|l|}{ACSCP~\cite{deep-gan-loss-ACSCP-18}}           & CVPR                  & 2018 & 75.7             & \textbf{102.7}           & 17.2            & 27.4             & 291.0           & 404.6          & -             & -             \\
\multicolumn{1}{|l|}{IG-CNN~\cite{deep-auto-IGCNN-18}}          & CVPR                  & 2018 & 72.5             & 118.2           & 13.6            & 21.1             & 291.4           & 349.4          & -             & -             \\
\multicolumn{1}{|l|}{Deep-NCL~\cite{deep-training-ConvNet-18}}        & CVPR                  & 2018 & 73.5             & 112.3           & 18.7            & 26.0             & 288.4           & 404.7          & -             & -             \\ \hline
\multicolumn{1}{|l|}{MCNN~\cite{MCNN-16}}            & CVPR                  & 2016 & 110.2            & 173.2           & 26.4            & 41.3             & 377.6           & 509.1          & 1.07          & 1.35          \\
\multicolumn{1}{|l|}{CSRNet~\cite{CSRNet-18}}          & CVPR                  & 2018 & 68.2             & 115.0           & 10.6            & 16.0             & 266.1           & 397.5          & 1.16          & 1.47          \\
\multicolumn{1}{|l|}{ic-CNN~\cite{deep-training-ICCNN-18}}           & ECCV                  & 2018 & 68.5  & 116.2 & 10.7 & 16.0   & 260.9 & 365.5 & 1.14 & 1.43       \\ \hline
\multicolumn{1}{|l|}{MCNN+McML}     & \multicolumn{1}{c}{-} & \multicolumn{1}{c|}{-} & 101.5 & 157.7 & 19.8 & 33.9 & 311.0   & 402.4 & 1.03  & 1.24     \\
\multicolumn{1}{|l|}{CSRNet+McML}   & \multicolumn{1}{c}{-} & \multicolumn{1}{c|}{-} & \textbf{59.1}  & 104.3 & \textbf{8.1}  & \textbf{10.6} & 246.1 & 367.7 & 1.01 & 1.27          \\
\multicolumn{1}{|l|}{ic-CNN+McML}    & \multicolumn{1}{c}{-} & \multicolumn{1}{c|}{-}  & 63.8  & 110.5 & 10.1 & 13.9 & \textbf{242.9} & 357.0   & \textbf{1.00}    & \textbf{1.20}   \\ \hline
\end{tabular}
\end{center}
\vspace{-2mm}
\label{tab:stoa-1}
\end{table*}

\vspace{-2mm}
\subsection{Ablation Studies}
\label{sec:ablation-studies}
We have conduct extensive ablation studies on our McML.

\noindent\textbf{MIE vs. MLS}.~We separately investigate the roles of our proposed two improvements,
\ljx{i.e.,~Mutual Learning Scheme~(MLS) and Mutual Information Estimation~(MIE).}
Experimental results are shown in Table~\ref{tab:ablation-study}.
Org. is the original baseline, MLS means that we ignore the mutual information estimation~(i.e.,~$\widehat{\II}\ljx{_\omega}$) in \Eqs~\ref{eq:column-1} and \ref{eq:column-2}, and MIE indicates that we optimize all columns at the same time~(i.e.,~do not alternately optimize each column). 
Generally speaking, MLS achieves better performance than all original baselines. 
After integrated MIE, there is a noticeable improvement.
It fully demonstrates the effectiveness of our method.

\noindent\textbf{Statistical Network}.~We intend to compare different statistical networks.
We have modified the proposed statistical network as follows:
1)~Only-1-Conv means only keep the last convolutional layer.
2)~Last-3-Conv denotes to preserve the last three convolutional layers.
3)~First-3-Conv indicates to retain the first three convolutional layers.
4)~FC-3 (64) means to add one fully connected \ljx{(FC)} layer with 64 outputs between the original \ljx{two FC layers. }
5)~FC-1 (64) indicates to reduce the outputs of the \ljx{first FC layer} into 64. 
6)~FC-1 (256) states to increase the outputs of the \ljx{first FC layer} into 256.
Comparison results are illustrated in Table \ref{tab:ablation-statistical-network}.
In general, different statistical networks have no significant difference in performance. 
Even using only one convolutional layer, our proposed training strategy still obviously improve the original baseline.
These results fully demonstrate the robustness of our method.

\begin{figure}[!b]
\begin{minipage}[b]{0.5\linewidth} 
\centering 
\includegraphics[width=0.9\linewidth]{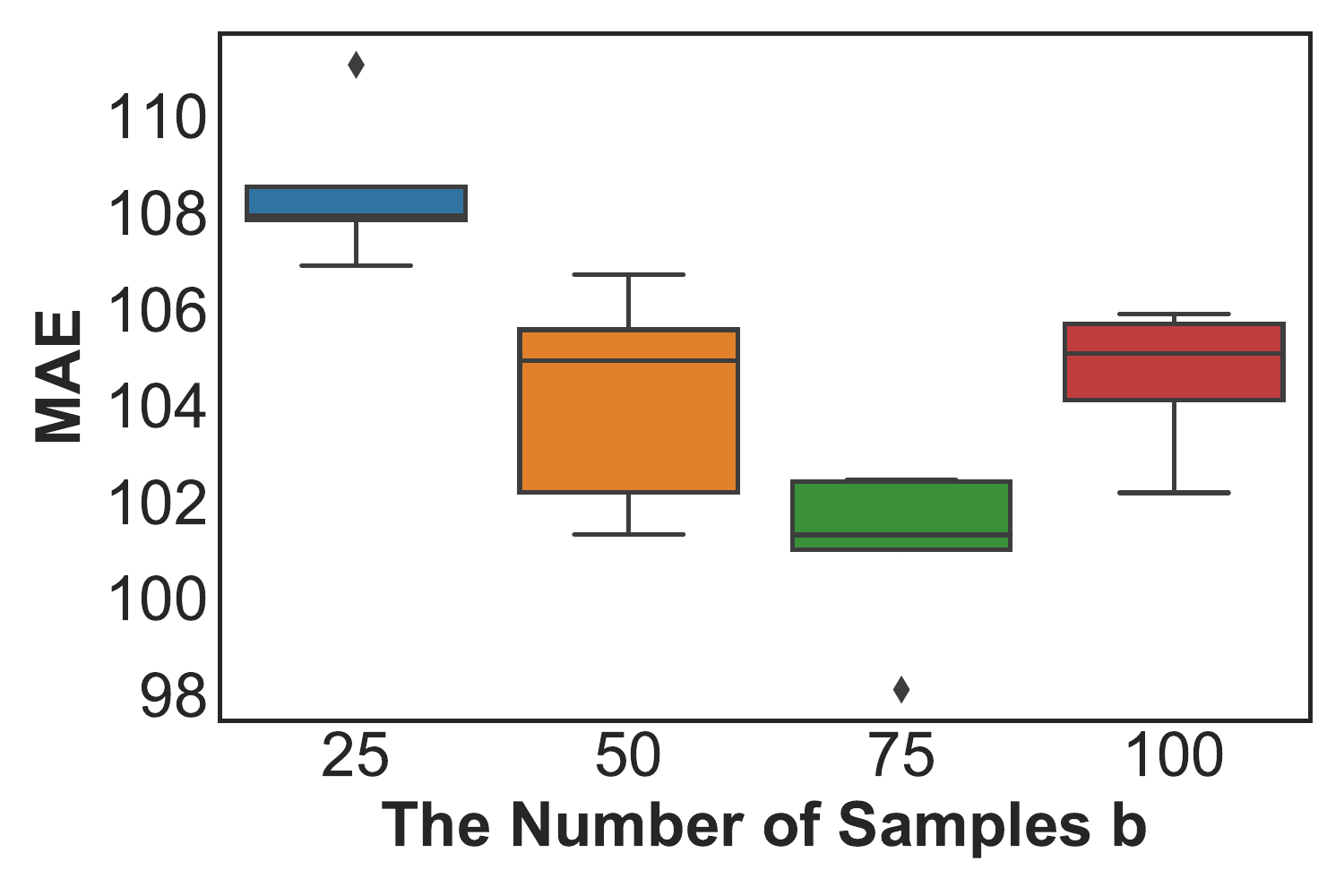}
\caption{Effects of samples $b$.} 
\label{fig:value-b} 
\end{minipage}%
\hspace{.1in}
\begin{minipage}[b]{0.45\linewidth} 
\normalsize
\centering
\begin{tabular}{|l|c|}
\hline
\textbf{Datasets}                  & $\mathbf{\alpha}$  \\ \hline \hline
ShanghaiTech A            & 0.3         \\
ShanghaiTech B            & 0.2         \\
UCF\_CC\_50 & 0.4         \\
UCSD    & 0.1         \\
WorldExpo'10              & 0.2         \\
\hline
\end{tabular}
\tabcaption{The values of $\alpha$.} 
\label{tab:value-a} 
\end{minipage} 
\end{figure}

\noindent\textbf{The number of samples $\mathbf{b}$}.~We study the effect of the number of samples $b$.
As shown in Figure \ref{fig:value-b}, we obverse that with the number of $b$ increases, the performance first increases and then decreases. 
Typically, when $b$ is too small, because of the estimated mutual information has a severe bias, our method intuitively gets poor performance.
\czq{In contrast}, when $b$ is too large, although the mutual information has been accurately estimated, the performance of our model is still severely affected since the iterations of the mutual learning \ljx{scheme} are inadequate.
Based on that we use a binary search to find the best value of $b$.
After extensive cross-validation, $b$ is set to 75 for all baselines.

\noindent\textbf{The weight of $\mathbf{\alpha}$}.~We have verified the impact of the weight of $\alpha$. 
To get a more accurate setting, we perform a grid search with the step of 0.1. 
The best values of $\alpha$ for different datasets are illustated in Table~\ref{tab:value-a}. 
Since ShanghaiTech Part A and UCF\_CC\_50 have more substantial scale changes, they have a larger $\alpha$ than other datasets.
We assume that the weight of $\alpha$ positively correlates to the degree of scale changes.

\vspace{-2mm}
\subsection{Comparisons with State-of-the-art}
\label{sec:compare-with-sota}
We demonstrate the efficiency of our McML on four challenging crowd counting datasets.
Tables~\ref{tab:stoa-1} and \ref{tab:sota-2} show the comparison with the other state-of-the-art methods.
We observe that McML can significantly improve three baselines~(i.e., MCNN, CSRNet, and ic-CNN) on all datasets. 
Notably, after using McML, the optimized CSRNet and ic-CNN also obviously outperform the other state-of-the-art approaches.
It fully demonstrates that our method can not only be applied to any \ljx{m}ulti-column network but also works on both dense and sparse crowd scenes.
Additionally, although ic-CNN also propose an alternate training process, our McML can still achieve better results than the original ic-CNN. 
It means that our McML is more effective than ic-CNN.

\begin{table}[!t]
\small
\renewcommand\arraystretch{0.9}
\caption{\small Ablation studies of statistical networks on ShanghaiTech Part A dataset~\cite{MCNN-16}.}
\label{tab:ablation-statistical-network}
\vspace{-2mm}
\setlength\tabcolsep{4pt}
\begin{tabular}{|l|c|c|c|c|c|c|}
\hline
 & \multicolumn{2}{c|}{\textbf{MCNN~\cite{MCNN-16}}} & \multicolumn{2}{c|}{\textbf{CSRNet~\cite{CSRNet-18}}} & \multicolumn{2}{c|}{\textbf{ic-CNN~\cite{deep-training-ICCNN-18}}} \\ \hline
\textbf{Structures} & \textbf{MAE}     & \textbf{MSE}    & \textbf{MAE}      & \textbf{MSE}     & \textbf{MAE}      & \textbf{MSE}     \\ \hline \hline
Only-1-Conv    & 104.2          & 160.8          & 61.7          & 106.9          & 66.2          & 114.1           \\
Last-3-Conv    & 103.5          & 160.1          & 61.1          & 106.8          & 65.5          & 113.3           \\
First-3-Conv   & 103.2          & 159.7          & 60.7          & 106.1          & 64.8          & 113.2           \\
\ljx{FC-3 (64)}      & 101.6          & 157.8          & 59.3          & \textbf{104.3} & 63.9          & \textbf{110.5}  \\
\ljx{FC-1 (64)}    & 102.0          & 158.2          & 59.8          & 105.1   & 64.5          & 111.3           \\
\ljx{FC-1 (256)}          & 102.2          & 158.4          & 59.7          & 104.8          & 63.9          & 111.0 \\
\ljx{Ours (Table~\ref{tab:statistics-network})}        & \textbf{101.5} & \textbf{157.7} & \textbf{59.1} & \textbf{104.3} & \textbf{63.8} & \textbf{110.5} \\
\hline
\end{tabular}
\end{table}

\begin{table}
\small
\caption{\small Comparison with state-of-the-art methods on WorldExpo'10~\cite{r-count-privacy-preserving-08} dataset. Only MAE is computed for each scene and then averaged to evaluate the overall performance.}
\vspace{-2mm}
\renewcommand\arraystretch{0.95}
\setlength\tabcolsep{5pt}
\begin{center}
\begin{tabular}{|l|c|c|c|c|c|c|}
\hline
\textbf{Method}        & \textbf{S1} & \textbf{S2} & \textbf{S3} & \textbf{S4} & \textbf{S5} & \textbf{Avg.} \\ \hline
Zhang et al.~\cite{deep-crowd-scene-15}  & 9.8    & 14.1   & 14.3   & 22.2   & 3.7    & 12.9     \\
Huang et al.~\cite{deep-body-BSAD-18}  & 4.1    & 21.7   & 11.9   & 11.0   & 3.5    & 10.5     \\
Switch-CNN~\cite{deep-switch-17}    & 4.4    & 15.7   & 10.0   & 11.0   & 5.9    & 9.4      \\
SaCNN~\cite{deep-fscale-multitask-SaCNN-18}    & \textbf{2.6} & 13.5    & 10.6   & 12.5  & \textbf{3.3}  & 8.5          \\
CP-CNN~\cite{deep-multi-pyramid-cnn-17}        & 2.9    & 14.7   & 10.5   & {10.4}   & 5.8    & 8.9      \\ \hline
MCNN~\cite{MCNN-16}          & 3.4    & 20.6   & 12.9   & 13.0   & 8.1    & 11.6     \\ 
CSRNet~\cite{CSRNet-18}        & 2.9    & {11.5}   & 8.6    & 16.6   & 3.4    & 8.6      \\ 
ic-CNN~\cite{deep-training-ICCNN-18} & 17.0 & 12.3 & 9.2  & 8.1  & 4.7 & 10.3  \\ \hline
MCNN+McML   & 3.4  & 15.2 & 14.6 & 12.7 & 5.2 & 10.2  \\
CSRNet+McML  & 2.8  & \textbf{11.2} & 9.0   & 13.5   & 3.5  & \textbf{8.0}  \\
ic-CNN+McML  & 10.7   & \textbf{11.2} & \textbf{8.2} & \textbf{8.0} & 4.5    & 8.5    \\ \hline
\end{tabular}
\end{center}
\vspace{-2mm}
\label{tab:sota-2}
\end{table}

For ShanghaiTech dataset, McML significantly boosts MCNN, CSRNet, and ic-CNN with relative MAE improvements of 7.9\%, 13.3\% and 6.9\% on Part A, and 25.0\%, 23.6\% and 5.6\% on Part B, respectively.
Similarly, for UCF\_CC\_50 dataset, McML provides the relative MAE improvements of 17.6\%, 7.5\%, and 6.9\% for three baselines.
These results clearly state McML can not only handle dense-crowd scenes but also work for small datasets.
On the other hand, experimental results of UCSD dataset show McML can improve the accuracy (i.e.,~lower MAE) and gain the robustness (i.e.,~lower MSE).
This result states the effectiveness of McML on the sparse-crowd scene.
Additionally, on WorldExpo'10 dataset, although our proposed McML \czq{does} not utilize perspective maps, they still achieve better results than other state-of-the-art methods that use perspective maps.

\begin{figure*}
\small
\begin{center}
\includegraphics[width=0.97\textwidth,height=0.23\textheight]{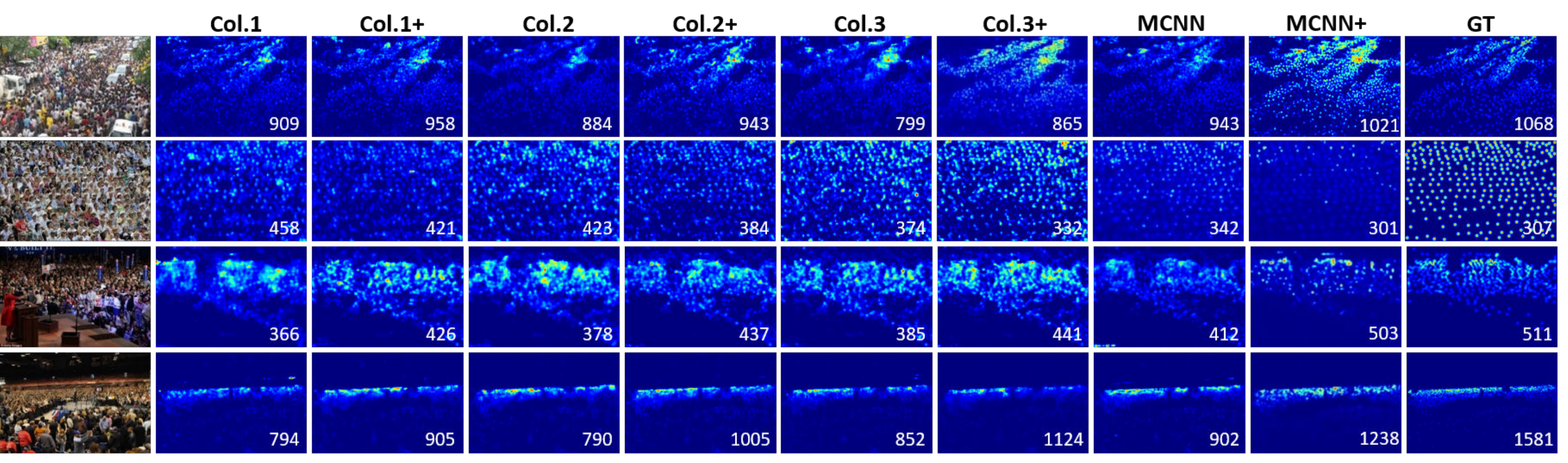}
\end{center}
\vspace{-4mm}
\caption{\small Comparison of estimated density maps between MCNN~\cite{MCNN-16} and McML.  
`+' indicates employing McML on the original MCNN.}
\label{fig:result}
\vspace{-2mm}
\end{figure*}

\vspace{-4mm}
\subsection{Why does McML Work}
\label{sec:how-ana-why-works}
We attempt to give more insights to show why our McML works.
The statistical analysis is illustrated in Table~\ref{tab:why}.
Compared with the results without McML (in Table~\ref{tab:Multi-columns-problem}), we observe that McML can significantly reduce Maximal Information Coefficient (MIC) and Structural SIMilarity (SSIM) between columns.
It denotes that our method can indeed reduce the \czq{redundant} parameters of columns and avoid overfitting.
\czq{On the other hand, McML can efficiently improve MIC and SSIM between the ensemble of all columns and the ground truth.}
It means that our method can guide multi-column structures to learn different scale features and improve the accuracy of crowd counting.

\begin{table}[t]
\small
\caption{\small The \czq{result} analysis of our proposed McML.
The values in the table are the average of all columns.
Col.$\leftrightarrow$Col. is the result between different columns.
Col.$\leftrightarrow$GT is the result between the ensemble of all columns and the ground truth.}
\renewcommand\arraystretch{0.9}
\vspace{-2mm}
\begin{center}
\begin{tabular}{|l|c|c|c|c|} 
\hline
& \multicolumn{2}{c|}{\textbf{Col.$\leftrightarrow$ Col. }}      & \multicolumn{2}{c|}{\textbf{Col.$\leftrightarrow$ GT }}  \\ 
\hline
\textbf{Method }  & \textbf{MIC } & \textbf{SSIM}  & \textbf{MIC } & \textbf{SSIM}           \\ 
\hline\hline
\multicolumn{5}{|c|}{ShanghaiTech Part A~\cite{MCNN-16}}             \\ 
\hline
MCNN+McML     &  0.74   & 0.61   &   0.68  &     0.70  \\
CSRNet+McML    & 0.77   & 0.70   &  0.82   &   0.82    \\
ic-CNN+McML    &  0.76   & 0.55    &   0.80  &     0.76  \\ 
\hline\hline
\multicolumn{5}{|c|}{{UCF\_CC\_50}~\cite{r-mrf-count-multi-source-13}}             \\ 
\hline
MCNN+McML     &  0.69   &  0.48  &  0.79   & 0.47      \\
CSRNet+McML     &  0.73   & 0.60  &  0.75   &  0.61     \\
ic-CNN+McML    &  0.75   & 0.60  &  0.72   &   0.64    \\
\hline
\end{tabular}
\end{center}
\vspace{-2mm}
\label{tab:why}
\end{table}

\czq{To further verify that our McML can indeed guide multi-column networks to learn different scales, we showcase the generated density maps from different columns of MCNN in Figure \ref{fig:result}.}
These four examples typically contain different crowd densities, occlusions, and scale changes.
\czq{We observe that estimated density maps of McML have more different salient areas than the original MCNN.}
It means that our method can indeed guide multi-column structures to focus on different scale information (i.e.,~different people/head sizes).
It is noted that the ground truth itself is generated with center points of pedestrians' heads, which inherently contains inaccurate information. 
Thus the result of our method is still unable to produce the same density map to the ground truth.

\vspace{-1mm}
\section{Conclusion}
\label{sec:conclusion}
In this paper, we propose a novel learning strategy called Multi-column Mutual learning (McML) for crowd counting, which can improve the scale invariance of feature learning and reduce parameter redundancy to avoid overfitting.
It could be applied to all existing CNN-based multi-column networks and is end-to-end trainable. 
Experiments on four challenging datasets fully demonstrate that it can significantly improve all baselines and outperforms the other state-of-the-art methods.
\czq{In summary, this} work provides the elegant views of effectively using multi-column architectures to improve the scale invariance.
In future work, we will study how to handle different image scales and resolutions in the ground truth generation.

\vspace{-1mm}
\section{ACKNOWLEDGEMENTS}

{\small This research was supported in part through the financial assistance award 60NANB17D156 from U.S. Department of Commerce, National Institute of Standards and Technology and by the Intelligence Advanced Research Projects Activity (IARPA) via Department of Interior/Interior Business Center (DOI/IBC) contract number D17PC00340, National Natural Science Foundation of China (Grant No: 61772436), Foundation for Department of Transportation of Henan Province, China (2019J-2-2), Sichuan Science and Technology Innovation Seedling Fund (2017RZ0015), China Scholarship Council (Grant No. 201707000083) and Cultivation Program for the Excellent Doctoral Dissertation of Southwest Jiaotong University (Grant No. D-YB 201707).}

\clearpage

\bibliographystyle{ACM-Reference-Format}
\bibliography{ms}

\end{document}